%% file: icdm_arxiv.tex
\documentclass[10pt, conference, compsocconf]{IEEEtran}
\ifCLASSINFOpdf
\else
\fi

\usepackage{amsmath}

\usepackage{multirow,bigdelim}

\usepackage{mathtools}
\usepackage{amssymb}

\usepackage{url}

\usepackage{tikz}
\usetikzlibrary{bayesnet}

 \usepackage{refcount}

\usepackage{hhline}

\usepackage{subcaption}

\usepackage[colorinlistoftodos,prependcaption,textsize=tiny]{todonotes}

\DeclarePairedDelimiterX{\expectarg}[1]{[}{]}{%
  \ifnum\currentgrouptype=16 \else\begingroup\fi
  \activatebar#1
  \ifnum\currentgrouptype=16 \else\endgroup\fi
}
\newcommand{\prob}{\operatorname{Pr}\probarg}
\DeclarePairedDelimiterX{\probarg}[1]{(}{)}{%
  \ifnum\currentgrouptype=16 \else\begingroup\fi
  \activatebar#1
  \ifnum\currentgrouptype=16 \else\endgroup\fi
}
\newcommand{\innermid}{\nonscript\;\delimsize\vert\nonscript\;}
\newcommand{\activatebar}{%
  \begingroup\lccode`\~=`\|
  \lowercase{\endgroup\let~}\innermid
  \mathcode`|=\string"8000
}
\newcommand{\matr}[1]{\mathbf{#1}}
\RequirePackage{xcolor}

\newcommand{\specialcell}[2][c]{%
  \begin{tabular}[#1]{@{}c@{}}#2\end{tabular}}
  
\renewcommand{\vec}{\boldsymbol}

\begin{document}
%
\title{MetaLDA: a Topic Model that Efficiently Incorporates Meta information}




%
\author{\IEEEauthorblockN{He Zhao\IEEEauthorrefmark{1},
Lan Du\IEEEauthorrefmark{1},
Wray Buntine\IEEEauthorrefmark{1} and 
Gang Liu\IEEEauthorrefmark{2}} 
\IEEEauthorblockA{\IEEEauthorrefmark{1}Faculty of Information Technology\\
Monash University,
Melbourne VIC, Australia\\ Email: \{he.zhao, lan.du, wray.buntine\}@monash.edu}
\IEEEauthorblockA{\IEEEauthorrefmark{2}College of Computer Science and Technology\\ 
Harbin Engineering University, Harbin, China\\
Email: liugang@hrbeu.edu.cn}}


\maketitle

\begin{abstract}
Besides the text content, documents and their associated words usually come with rich sets of meta information, 
such as categories of documents and semantic/syntactic features of words, like those encoded in word embeddings. 
Incorporating such meta information directly into the generative process of topic models can improve modelling accuracy and topic quality,
especially in the case where the word-occurrence information in the training data is insufficient. 
In this paper, we present a topic model, called MetaLDA, which is able to leverage either document or word meta information, or both of them jointly.
With two data argumentation techniques, we can 
derive an efficient Gibbs sampling algorithm, which benefits from the fully local conjugacy of the model.
Moreover, the algorithm is favoured by the sparsity of the meta information.
Extensive experiments on several real world datasets demonstrate that our 
model achieves comparable or improved performance in terms of both perplexity and topic quality,
particularly in handling sparse texts. 
In addition, compared with other models using meta information, 
our model runs significantly faster. 

\end{abstract}

\begin{IEEEkeywords}
topic models; meta information; short texts;

\end{IEEEkeywords}

%
\IEEEpeerreviewmaketitle

\section{Introduction}

With the rapid growth of the internet, huge amounts of text data are generated
in social networks, online shopping and news websites, etc. 
These data
create demand for powerful and efficient text analysis techniques.  
Probabilistic topic models such as Latent Dirichlet Allocation
(LDA)~\cite{blei2003latent} are popular approaches for this task, 
by discovering latent topics from text collections. 
Many conventional topic models discover topics purely based on the word-occurrences, 
ignoring the \textit{meta information} (a.k.a., \textit{side information}) associated with the content. 
In contrast, when we humans read text
it is natural to leverage meta information to improve our comprehension, 
which includes categories, authors, timestamps, the semantic meanings of the words, etc. 
Therefore, topic models capable of using meta information should yield improved modelling accuracy and topic quality.

In practice, various kinds of meta information are available at the document level and the word level in many corpora. 
At the document level, 
labels of documents can be used to guide topic learning so that more meaningful topics 
can be discovered.
Moreover, it is highly likely that documents  with
common labels discuss similar topics, which could further result in similar topic distributions. 
For example, if we use authors as labels for scientific papers, the topics of the
papers published by the same researcher can be closely related.

At the word level, different semantic/syntactic features are also accessible. 
For example, there are features regarding word relationships, such as synonyms obtained
from WordNet~\cite{miller1995wordnet}, word co-occurrence patterns  obtained from a large
corpus, and linked concepts from knowledge graphs. 
It is preferable that words having similar meaning but different morphological forms, 
like ``dog'' and ``puppy'', are assigned to the same topic, even if they barely co-occur in the modelled corpus.  
Recently, word
embeddings generated by GloVe~\cite{pennington2014glove} and
word2vec~\cite{mikolov2013distributed}, have attracted a lot of attention
in natural language processing and related fields.
It has been shown that the word embeddings can capture both the semantic and syntactic features of words so 
that similar words are close to each other
in the embedding space.
It seems reasonable to expect that these word embedding will
improve topic modelling \cite{das2015gaussian,nguyen2015improving}. 

Conventional topic models can suffer from a large performance degradation over short
texts (e.g., tweets and news headlines) because of insufficient word co-occurrence
information. In such cases, meta information of documents and words can
play an important role in analysing short texts by compensating 
the lost information in word co-occurrences. 
At the document level, for
example, tweets are usually associated with hashtags, users, locations, and
timestamps, which can be used to alleviate the data sparsity problem. At the word level, word semantic similarity and embeddings obtained or trained on large external corpus (e.g.,
Google News or Wikipedia) have been proven useful in learning meaningful topics from
short texts \cite{xun2016topic,li2016topic}.

The benefit of using document and word meta information separately is
shown in several models such as \cite{mimno2012topic,ramage2011partially,nguyen2015improving}. 
However, in existing models this is usually 
not efficient enough due to non-conjugacy and/or complex model structures. 
Moreover, only one kind of meta information (either at document level or at word level) is used 
in most existing models. In this paper, we propose 
MetaLDA\footnote{Code at \url{https://github.com/ethanhezhao/MetaLDA/}}, 
a topic model
that can effectively and efficiently leverage arbitrary document and word meta
information encoded in binary form. 
Specifically, the labels of a document in
MetaLDA are incorporated in the prior of the per-document topic distributions. 
If two documents have similar labels,  their topic distributions should be generated with similar Dirichlet priors. 
Analogously, at the word level, the features of a word are incorporated in the
prior of the per-topic word distributions, which encourages words with similar
features to have similar weights across topics. 
Therefore, both document and word meta information,
if and when they are available, can be flexibly and simultaneously 
incorporated using MetaLDA.
%
MetaLDA has the following key properties: 
\begin{enumerate}     
  \item MetaLDA jointly
incorporates various kinds of document and word meta information for both
regular and short texts, yielding better modelling accuracy and topic quality.   
  \item With the data augmentation techniques, the inference of MetaLDA can be done by an efficient and
closed-form Gibbs sampling algorithm that benefits from the full local conjugacy of the model. 
  \item The simple structure of incorporating meta information and the efficient inference algorithm give MetaLDA advantage in terms of running speed over other models with meta information. 
\end{enumerate}

We conduct extensive experiments with several real datasets including regular
and short texts in various domains.
The experimental results demonstrate that
MetaLDA achieves improved performance in terms of perplexity, topic
coherence, and running time.

\section{Related Work}
\label{section-related}
In this section, we review three lines of related work: models with document
meta information, models with word meta information, and models for short texts. 

At the document level, Supervised LDA
(sLDA)~\cite{mcauliffe2008supervised} models document labels by
learning a generalised linear model with an appropriate link function and
exponential family dispersion function. But the restriction for sLDA is that one
document can only have one label. Labelled LDA (LLDA)~\cite{ramage2009labeled}
assumes that each label has a corresponding topic and a document is generated by
a mixture of the topics. Although multiple labels
are allowed, LLDA requires that the number of topics must equal to the number of labels,
i.e., exactly one topic per label.
As an extension to LLDA, Partially Labelled LDA (PLLDA)~\cite{ramage2011partially}
relaxes this requirement by assigning multiple topics to a label. The Dirichlet
Multinomial Regression (DMR) model~\cite{mimno2012topic} incorporates document labels on the prior of
the topic distributions like our MetaLDA but with the logistic-normal transformation. As
full conjugacy does not exist in DMR, a part of the inference has to be done by
numerical optimisation, which is slow for large sets of labels and topics. Similarly,
in the Hierarchical Dirichlet Scaling Process (HDSP)~\cite{Kim2017}, conjugacy
is broken as well since the topic distributions have to be renormalised.
\cite{hu2016non} introduces a Poisson factorisation model with hierarchical
document labels. But the techniques cannot be applied to regular topic models
as the topic proportion vectors are also unnormalised.

Recently, there is growing interest in incorporating word features in topic
models. For example, DF-LDA~\cite{andrzejewski2009incorporating} incorporates
word must-links and cannot-links using a Dirichlet forest prior in LDA; MRF-LDA~\cite{xie2015incorporating} 
encodes word semantic similarity in LDA with a Markov
random field; WF-LDA~\cite{petterson2010word} extends LDA to model
word features with the logistic-normal transform; LF-LDA~\cite{nguyen2015improving} 
integrates word embeddings into LDA by replacing
the topic-word Dirichlet multinomial component with a mixture of a Dirichlet
multinomial component and a word embedding component; Instead of generating word
types (tokens), Gaussian LDA (GLDA)~\cite{das2015gaussian}
directly generates word embeddings with the Gaussian distribution. 
Despite the exciting applications of the above models, their
inference is usually less efficient due to the non-conjugacy and/or complicated
model structures.

Analysis of short text with topic models has been an
active area with the development of social networks.
Generally, there are two ways to deal with the sparsity problem in short texts, either using the intrinsic properties of short texts or 
leveraging meta information. 
For the first way, one popular approach is to aggregate short texts into pseudo-documents, 
for example, \cite{hong2010empirical} introduces a model that aggregates tweets containing the same word;
Recently, PTM~\cite{zuo2016topic} aggregates short texts into latent pseudo documents. 
Another approach is to assume one topic per short document, 
known as mixture of unigrams or Dirichlet Multinomial Mixture (DMM) such as~\cite{yin2014dirichlet,xun2016topic}. 
For the second way, document meta information can be used to aggregate short texts, 
for example, \cite{hong2010empirical} aggregates tweets by the corresponding authors and  
\cite{mehrotra2013improving} shows that aggregating tweets by their hashtags yields superior performance over 
other aggregation methods. 
One closely related work to ours is the models that use word features for short texts. 
For example, \cite{xun2016topic} introduces an extension of GLDA on short texts which samples
an indicator variable that chooses to generate either the type of a word or the embedding
of a word and GPU-DMM~\cite{li2016topic} extends DMM with word semantic similarity obtained from embeddings for short texts.
Although with improved performance there still exists challenges for existing models: 
(1) for aggregation-based models, it is usually hard to choose which meta information to use for aggregation; 
(2) the ``single topic'' assumption makes DMM models lose the flexibility to capture different topic ingredients of a document; 
and (3) the incorporation of meta information in the existing models is usually less efficient.

To our knowledge, the attempts that jointly leverage document and word meta
information are relatively rare. For example, meta information can be
incorporated by first-order logic in Logit-LDA~\cite{andrzejewski2011framework}
and score functions in SC-LDA~\cite{yang2015efficient}. However, the first-order
logic and score functions need to be defined for different kinds of meta information
 and the definition can be infeasible for incorporating both document and
word meta information simultaneously. 

\section{The MetaLDA Model}
\label{section-model}

Given a corpus, LDA uses the same Dirichlet prior for all the per-document topic distributions 
and the same prior for 
all the per-topic word distributions~\cite{wallach2009rethinking}. 
While in MetaLDA, each document has a specific Dirichlet prior on its topic distribution, which
is computed from the meta information of the document,
and the parameters of the prior are estimated during training.
Similarly, each topic has a specific Dirichlet prior computed from the word meta information.
%
Here we elaborate our MetaLDA, in particular on how
the meta information is incorporated. 
Hereafter, we will use labels as document meta information, unless otherwise stated.


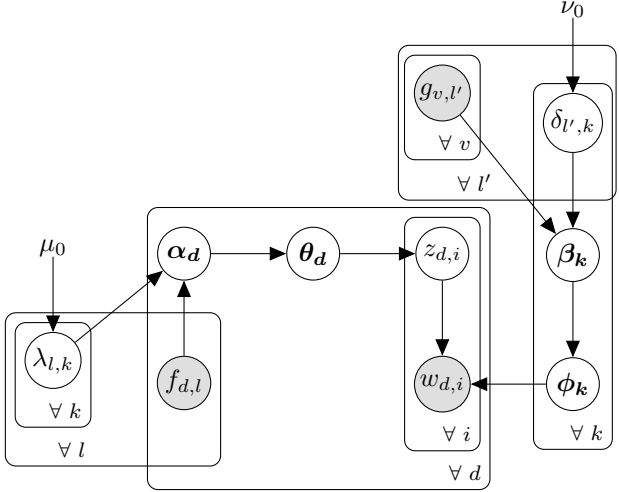
\begin{figure}[t!]
  \begin{center}
    \input{ExtLDA_graphical_model}
  \end{center}
  \caption{The graphical model of MetaLDA}
   \label{fig-model}
\end{figure}

Given a collection of $D$ documents
$\mathcal{D}$, 
MetaLDA generates document $d \in \{1,\cdots,D\}$ with a mixture
of $K$ topics and each topic $k \in \{1,\cdots,K\}$ is a distribution over the
vocabulary with $V$ tokens, denoted by $\vec{\phi}_{k} \in \mathbb{R}^{V}_{+}$.
For document $d$ with $N_d$ words, to generate the $i^{\text{th}}$ ($i \in
\{1,\cdots,N_d\}$) word $w_{d,i}$, we first sample a topic $z_{d,i} \in
\{1,\cdots,K\}$ from the document's topic distribution $\vec{\theta_d} \in
\mathbb{R}^{K}_{+}$, and then sample $w_{d,i}$ from $\vec{\phi}_{z_{d,i}}$.
Assume the labels of document $d$ are encoded in a binary vector $\vec{f_d} \in
\{0,1\}^{L_{doc}}$ where $L_{doc}$ is the total number of unique labels. $f_{d,l}=1$
indicates label $l$ is active in document $d$ and vice versa. Similarly, the
$L_{word}$ features of token $v$ are stored ∂in a binary vector $\vec{g}_v \in
\{0,1\}^{L_{word}}$. 
Therefore, the document and word meta information associated with $\mathcal{D}$ are 
stored in the matrix $\matr{F} \in \{0,1\}^{D \times L_{doc}}$ and $\matr{G} \in \{0,1\}^{V \times L_{word}}$ respectively.
Although MetaLDA incorporates binary features, categorical
features and real-valued features can be converted into binary values with
proper transformations such as discretisation and binarisation.

Fig.~\ref{fig-model} shows the graphical model of MetaLDA and the generative 
process is as following:
\begin{enumerate}
  \item  For each topic $k$:
    \begin{enumerate}
      \item For each doc-label $l$: Draw $\lambda_{l,k} \sim \mathrm{Ga}(\mu_0,\mu_0)$
      \item For each word-feat $l'$: Draw $\delta_{l',k} \sim \mathrm{Ga}(\nu_0,\nu_0)$
      \item \label{step_beta} For each token $v$: Compute $\beta_{k,v} = \prod_{l'=1}^{L_{word}} \delta_{l',k}^{g_{v,l'}}$
      \item Draw $\vec{\phi}_k \sim \text{Dir}_V(\vec{\beta}_{k})$
    \end{enumerate}
  \item For each document $d$:
    \begin{enumerate}
      \item \label{step_alpha} For each topic $k$: Compute $\alpha_{d,k} = \prod_{l=1}^{L_{doc}} \lambda_{l,k}^{f_{d,l}}$ 
      \item Draw $\vec{\theta}_d \sim \text{Dir}_K(\vec{\alpha}_d)$
      \item \label{step_word}For each word in document $d$:
      \begin{enumerate}
        \item Draw topic $z_{d,i} \sim \text{Cat}_K(\vec{\theta}_d)$
        \item Draw word $w_{d,i} \sim \text{Cat}_V(\vec{\phi}_{z_{d,i}})$
      \end{enumerate}
    \end{enumerate}
\end{enumerate} where $\text{Ga}(\cdot,\cdot)$, $\text{Dir}(\cdot)$,
$\text{Cat}(\cdot)$ are the gamma distribution, 
the  Dirichlet distribution, and the
categorical distribution respectively. $K$, $\mu_0$, and $\nu_0$ are the hyper-parameters.

To incorporate document labels, MetaLDA learns a specific
Dirichlet prior over the topics for each document by using the label information. 
Specifically, the information of
document $d$'s labels is incorporated in $\vec{\alpha}_d$, the parameter
of Dirichlet prior on $\vec{\theta}_d$.
As shown in Step~\ref{step_alpha}, $\alpha_{d,k}$ is
computed as a log linear combination of the labels $f_{d,l}$. 
Since $f_{d,l}$ is binary, $\alpha_{d,k}$ is indeed the multiplication of $\lambda_{l,k}$ over all
the active labels of document $d$, i.e., $\{l \mid f_{d, l}=1\}$. Drawn from the gamma
distribution with mean 1, $\lambda_{l,k}$ controls the impact of label $l$ on
topic $k$. If label $l$ has no or less impact on topic $k$, $\lambda_{l,k}$ is
expected to be 1 or close to 1, and then $\lambda_{l,k}$ will have no or little
influence on $\alpha_{d,k}$ and vice versa. The hyper-parameter $\mu_0$ controls
the variation of $\lambda_{l,k}$. The incorporation of word features is
analogous but in the parameter of the Dirichlet prior on the per-topic word distributions as shown in Step~\ref{step_beta}.

The intuition of our way of incorporating meta information is: At the
document level, if two documents have more labels in common, their Dirichlet
parameter $\vec{\alpha}_d$ will be more similar, resulting in more similar
topic distributions $\vec{\theta}_d$; At the word level, if two words have 
similar features, their $\beta_{k,v}$ in topic $k$ will be  similar and then
we can expect that their $\phi_{k,v}$ could be more or less the same. 
Finally, the two words will have similar probabilities of showing up in topic $k$. 
In other words, if a topic
``prefers'' a certain word,  we expect that it will also prefer other words with similar
features to that word.
Moreover, at both the document and the word level, different labels/features
may have different impact on the topics ($\lambda$/$\delta$), which is automatically learnt in MetaLDA.

\section{Inference}
Unlike most existing methods, our way of incorporating the meta information 
facilitates the derivation of
an efficient Gibbs sampling algorithm.
With two data augmentation techniques (i.e., the introduction of auxiliary variables),
MetaLDA admits the local conjugacy and a close-form Gibbs sampling algorithm can be derived.
Note that MetaLDA incorporates the meta information on the Dirichlet priors, so we can still use 
 LDA's collapsed Gibbs sampling algorithm for the topic assignment $z_{d,i}$.
 Moreover,
Step \ref{step_alpha} and \ref{step_beta} show that one only needs to consider the 
non-zero entries of $\matr{F}$ and $\matr{G}$ in computing the full conditionals,
which further reduces the inference complexity.
 

Similar to LDA, the complete model likelihood (i.e., joint distribution) of MetaLDA is:
\begin{IEEEeqnarray}{+rCl+x*}
\prod_{k=1}^{K} \prod_{v=1}^{V} \phi_{k,v}^{n_{k,v}} \cdot \prod_{d=1}^{D} \prod_{k=1}^{K} \theta_{d,k}^{m_{d,k}}
\label{eq-ll}
\end{IEEEeqnarray}
where $n_{k,v} = \sum_{d}^{D} \sum_{i=1}^{N_d} \boldsymbol{1}_{(w_{d,i} = v, z_{d,i} = k)}$, $m_{d,k} = \sum_{i=1}^{N_d} \boldsymbol{1}_{(z_{d,i} = k)}$, and $\boldsymbol{1}_{(\cdot)}$ is the indicator function.

\subsection{Sampling $\lambda_{l,k}$:}

To sample $\lambda_{l,k}$, we first marginalise out $\theta_{d,k}$  in the right part of Eq.~(\ref{eq-ll}) with the Dirichlet multinomial conjugacy:
\begin{IEEEeqnarray}{+rCl+x*}
 \prod_{d=1}^{D} \underbrace{\frac{\Gamma(\alpha_{d,\cdot})}{\Gamma(\alpha_{d,\cdot} + m_{d,\cdot})}}_{\text{Gamma ratio 1}} \prod_{k=1}^{K} \underbrace{\frac{\Gamma(\alpha_{d,k} + m_{d,k})}{\Gamma(\alpha_{d,k})}}_{\text{Gamma ratio 2}}
\label{eq-gamma-ratio}
\end{IEEEeqnarray}
where 
$\alpha_{d,\cdot} = \sum_{k=1}^{K} \alpha_{d,k}$, $m_{d,\cdot} = \sum_{k=1}^{K} m_{d,k}$, and $\Gamma(\cdot)$ is the gamma function.
Gamma ratio 1 in Eq.~(\ref{eq-gamma-ratio}) can be augmented with a set of Beta random variables $q_{1:D}$ as:
\begin{IEEEeqnarray}{+rCl+x*}
\underbrace{\frac{\Gamma(\alpha_{d,\cdot})}{\Gamma(\alpha_{d,\cdot} + m_{d,\cdot})}}_{\text{Gamma ratio 1}} & \propto &\int_{q_d} q_d^{\alpha_{d,\cdot}-1} (1-q_d)^{m_{d,\cdot}-1} 
\label{eq-q-d}
\end{IEEEeqnarray}
where for each document $d$, $q_d \sim \text{Beta}(\alpha_{d,\cdot}, m_{d,\cdot})$. 
Given a set of $q_{1:D}$ for all the documents, Gamma ratio 1 can be approximated
by the product of $q_{1:D}$, i.e., $\prod_{d=1}^{D} q_d^{\alpha_{d, \cdot}}$.

Gamma ratio 2 in Eq.~(\ref{eq-gamma-ratio}) is the Pochhammer symbol for a rising factorial, which can be augmented
with an auxiliary variable $t_{d,k}$~\cite{chen2011sampling,teh2012hierarchical,zhou2015negative,pmlr-v70-zhao17a} as follows:
\begin{IEEEeqnarray}{+rCl+x*}
\underbrace{\frac{\Gamma(\alpha_{d,k} + m_{d,k})}{\Gamma(\alpha_{d,k})}}_{\text{Gamma ratio 2}} &=& \sum_{t_{d,k}=0}^{m_{d,k}} S^{m_{d,k}}_{t_{d,k}} \alpha_{d,k}^{t_{d,k}}
\label{eq-stirling}
\end{IEEEeqnarray}
where $S^{m}_{t}$ indicates an unsigned Stirling number of the first kind.
Gamma ratio 2 is a normalising constant
for the probability of the number of tables in the Chinese Restaurant
Process (CRP)~\cite{BunHut:12}, $t_{d,k}$ can be sampled by a CRP with $\alpha_{d,k}$ as the concentration and $m_{d,k}$ as the number of customers:
\begin{IEEEeqnarray}{+rCl+x*}
t_{d,k} &=& \sum_{i=1}^{m_{d,k}} \text{Bern}\left(\frac{\alpha_{d,k}}{\alpha_{d,k}+i}\right)
\label{eq-crp}
\end{IEEEeqnarray} where $\text{Bern}(\cdot)$ samples from the Bernoulli
distribution. The complexity of sampling $t_{d,k}$ by Eq.~(\ref{eq-crp}) is $\mathcal{O}(m_{d,k})$. For large $m_{d,k}$, as the standard deviation of
$t_{d,k}$ is $\mathcal{O}(\sqrt{\log m_{d,k}})$ \cite{BunHut:12}, one can sample $t_{d,k}$ in a small window around the current value
in complexity $\mathcal{O}(\sqrt{\log m_{d,k}})$.

By ignoring the terms unrelated to $\alpha$, the augmentation of Eq.~(\ref{eq-stirling}) can be simplified to a single term $\alpha_{d,k}^{t_{d,k}}$.  With auxiliary variables now introduced, we simplify Eq.~(\ref{eq-gamma-ratio}) to:
\begin{IEEEeqnarray}{+rCl+x*}
\prod_{d=1}^{D} \prod_{k=1}^{K} q_d^{\alpha_{d,k}} \alpha_{d,k}^{t_{d,k}} 
\label{eq-alpha}
\end{IEEEeqnarray}
\noindent
Replacing $\alpha_{d,k}$ with $\lambda_{l,k}$, we can get:
\begin{IEEEeqnarray}{+rCl+x*}
\label{eq-lambda-1}
\prod_{d=1}^{D} \prod_{k=1}^{K} e^{- \alpha_{d,k} \log{\frac{1}{q_d}}} \cdot \prod_{l=1}^{L_{doc}} \prod_{k=1}^{K} \lambda_{l,k}^{\sum_{d=1}^{D} f_{d,l} t_{d,k} } \nonumber
\end{IEEEeqnarray}

Recall that all the document labels are binary and $\lambda_{l,k}$ is involved in computing $\alpha_{d,k}$ iff $f_{d,l}=1$. 
Extracting all the terms related to $\lambda_{l,k}$ in Eq.~(\ref{eq-lambda-1}), we get the marginal posterior of $\lambda_{l,k}$:
\begin{IEEEeqnarray}{+rCl+x*}
e^{- \lambda_{l,k}  \sum_{d=1:f_{d,l}=1}^{D} \log\frac{1}{q_d} \cdot \frac{\alpha_{d,k}}{\lambda_{l,k}}} \lambda_{l,k} ^{\sum_{d=1}^{D} f_{d,l} t_{d,k}}  \nonumber
\end{IEEEeqnarray}
where $\frac{\alpha_{d,k}}{\lambda_{l,k}}$ is the value of $\alpha_{d,k}$ with $\lambda_{l,k}$ removed when $f_{d,l}=1$. 
With the data augmentation techniques, the posterior is transformed into a form that is conjugate to the gamma prior of $\lambda_{l,k}$.
Therefore, it is straightforward to yield the following sampling strategy for $\lambda_{l,k}$:
\setlength{\belowdisplayskip}{9pt} \setlength{\belowdisplayshortskip}{9pt}
\begin{IEEEeqnarray}{+rCl+x*}
\label{eq-lambda-sample-1}
\lambda_{l,k} &\sim& \mathrm{Ga}( \mu', 1/\mu'')\\
\label{eq-lambda-sample-2}
\mu' &=& \mu_0 +  \sum_{d=1: f_{d,l} = 1}^{D}  t_{d,k} \\
\label{eq-lambda-sample-3}
\mu'' &=& 1/\mu_0 -  \sum_{d=1:f_{d,l}=1}^{D} \frac{\alpha_{d,k}}{\lambda_{l,k}} \log q_d
\end{IEEEeqnarray}

We can compute and cache the value of $\alpha_{d,k}$ first. After $\lambda_{l,k}$ is sampled, $\alpha_{d,k}$ can be updated by:
\setlength{\belowdisplayskip}{1pt} \setlength{\belowdisplayshortskip}{1pt}
\begin{IEEEeqnarray}{+rCl+x*}
\label{eq-alpha-update}
\alpha_{d,k} \leftarrow \frac{\alpha_{d,k} \lambda'_{l,k}}{\lambda_{l,k}} 
~\hfill \forall~1 \leq d \leq D : f_{d,l} = 1
\end{IEEEeqnarray}
where $\lambda'_{i,k}$ is the newly-sampled value of $\lambda_{i,k}$.
 
To sample/compute Eqs.~(\ref{eq-lambda-sample-1})-(\ref{eq-alpha-update}), one only iterates over the documents where label $l$ is active
(i.e., $f_{d,l}=1$). Thus, the sampling for all $\lambda$ takes
$\mathcal{O}(D'KL_{doc})$ where $D'$ is the average number of documents 
where a label is active (i.e., the column-wise sparsity of $\matr{F}$). It is usually that $D' \ll D$ because if a label exists
in nearly all the documents, it provides little discriminative information. This demonstrates
how the sparsity of document meta information is leveraged. Moreover,
sampling all the tables $t$ takes
$\mathcal{O}(\tilde{N})$ ($\tilde{N}$ is the total number of words in
$\mathcal{D}$) which can be accelerated with the window sampling technique
explained above.

\subsection{Sampling $\delta_{l',k}$:}

Since the derivation of sampling $\delta_{l',k}$ is analogous to $\lambda_{l,k}$, we directly give the sampling formulas:
\begin{IEEEeqnarray}{+rCl+x*}
\label{eq-delta-sample-1}
\delta_{l',k} &\sim& \mathrm{Ga}( \nu', 1/\nu'')\\
\label{eq-delta-sample-2}
\nu' &=& \nu_0 +  \sum_{v=1: g_{v,l'} = 1}^{V}  t'_{k,v} \\
\label{eq-delta-sample-3}
\nu'' &=& 1/\nu_0 -  \log q'_k \sum_{v=1:g_{v,l'}=1}^{V} \frac{\beta_{k,v}}{\delta_{l',k}}
\end{IEEEeqnarray}
where the two auxiliary variables can be sampled by: $q'_k \sim \text{Beta}(\beta_{k,\cdot}, n_{k,\cdot})$ and $t'_{k,v} \sim \text{CRP}(\beta_{k,v}, n_{k,v})$. Similarly, sampling all $\delta$ takes $\mathcal{O}(V'KL_{word})$ where $V'$ is the average number of tokens where a feature is active (i.e., the column-wise sparsity of $\matr{G}$ and usually $V' \ll V$) and sampling all the tables $t'$ takes $\mathcal{O}(\tilde{N})$.

\subsection{Sampling topic $z_{d,i}$:}
Given $\vec{\alpha_d}$ and $\vec{\beta_k}$, the collapsed Gibbs sampling of a new topic for a word $w_{d,i} = v$ in MetaLDA is:
\begin{IEEEeqnarray}{+rCl+x*}
\prob{z_{d,i} = k} &\propto& (\alpha_{d,k} + m_{d,k}) \frac{\beta_{k,v} + n_{k,v}}{\beta_{k,\cdot} + n_{k,\cdot}}
\end{IEEEeqnarray}
which is exactly the same to LDA.

\section{Experiments}

In this section, we evaluate the proposed MetaLDA against several recent advances that also incorporate meta information 
on 6 real datasets including both regular and short texts.
The goal of the experimental work is to evaluate the effectiveness and efficiency of MetaLDA's incorporation 
of document and word meta information both separately and
jointly compared with other methods. 
We report the performance in terms of perplexity, 
topic coherence, and running time per iteration.

\subsection{Datasets}

In the experiments, three regular text datasets and three short text datasets were used:
\begin{itemize}     
  \item \textbf{Reuters} is widely used corpus extracted from the Reuters-21578 dataset where documents without any labels are removed\footnote{\label{fn-pre-process} MetaLDA is able to handle documents/words without labels/features. 
  But for fair comparison with other models, we removed the documents without labels and words without features.}. 
  There are 11,367 documents and 120 labels. 
  Each document is associated with multiple labels.
  The vocabulary size is 8,817 and the average document length is 73.
  \item \textbf{20NG}, 20 Newsgroup, a widely used dataset consists of 18,846 news articles with 20 categories. 
  The vocabulary size is 22,636 and the average document length is 108.
  \item \textbf{NYT}, New York Times
  is extracted from the documents in the category ``Top/News/Health'' in the New York Times Annotated Corpus\footnote{https://catalog.ldc.upenn.edu/ldc2008t19}. 
  There are 52,521 documents and 545 unique labels.
  Each document is with multiple labels. The vocabulary contains 21,421 tokens and there are 442 words in a document on average. 
  \item \textbf{WS}, Web Snippet, used in \cite{li2016topic}, contains 12,237 web search snippets and each snippet belongs to one of 8 categories. The vocabulary contains 10,052 tokens and there are 15 words in one snippet on average. 
  \item \textbf{TMN}, Tag My News, used in \cite{nguyen2015improving},  consists of 32,597 English RSS news snippets from Tag My News. With a title and a short description, each snippet belongs to one of 7 categories. There are 13,370 tokens in the vocabulary and the average length of a snippet is 18.
  \item \textbf{AN}, ABC News, is a collection of 12,495 short news descriptions and each one is in multiple of 194 categories. There are 4,255 tokens in the vocabulary and the average length of a description is 13.
\end{itemize}

All the datasets were tokenised by Mallet\footnote{\url{http://mallet.cs.umass.edu}} and we removed the words that exist in less than 5 documents and more than 95\% documents. 

\subsection{Meta Information Settings}

\textbf{Document labels and word features.} At the document level, the labels associated with documents in each dataset were used as the meta information. 
At the word level, we used a set of 100-dimensional binarised word embeddings as word features\footnotemark[\getrefnumber{fn-pre-process}], 
which were obtained from the 50-dimensional GloVe word embeddings pre-trained on Wikipedia\footnote{\url{https://nlp.stanford.edu/projects/glove/}}. 
To binarise word embeddings, we first adopted the following method similar to \cite{guo2014revisiting}:
\begin{IEEEeqnarray}{+rCl+x*}
g'_{v,j} = 
\begin{cases}
1, & \text{if}\ g''_{v,j} > \text{Mean}_{+}(\vec{g}''_{v}) \\
-1, & \text{if}\ g''_{v,j} < \text{Mean}_{-}(\vec{g}''_{v}) \\
0, & \text{otherwise}
\end{cases}
\end{IEEEeqnarray}
where $\vec{g}''_{v}$ is the original embedding vector for word $v$, $g'_{v,j}$ is the binarised value for $j^{\text{th}}$ element of $\vec{g''_{v}}$, 
and $\text{Mean}_{+}(\cdot)$ and $\text{Mean}_{-}(\cdot)$ are the average value of all the positive elements and negative elements respectively. 
The insight is that we only consider features with strong opinions (i.e., large positive or negative value) on each dimension.
To transform $g' \in \{-1,1\}$ to the final $g \in \{0,1\}$, we use two binary bits to encode one dimension of $g'_{v,j}$: the first bit is on if $g'_{v,j} = 1$ 
and the second is on if $g'_{v,j} = -1$. 
Besides, MetaLDA can work with other word features such as semantic similarity as well.

\textbf{Default feature.} Besides the labels/features associated with the datasets, 
a default label/feature for each document/word is introduced in MetaLDA, 
which is always equal to 1. 
The default can be interpreted as the bias term in $\alpha$/$\beta$, which captures the information unrelated to the labels/features.
While there are no document labels or word features, with the default,
MetaLDA is equivalent in model
to asymmetric-asymmetric LDA of \cite{wallach2009rethinking}.


\begin{table}[]
\centering
\caption{MetaLDA and its variants.}
\label{table-variant}
\resizebox{0.48\textwidth}{!}{

\begin{tabular}{|c|c|c|}
\hline
                & Compute $\alpha$ with & Compute $\beta$ with       \\ \hline
MetaLDA         & Document labels       & Word features          \\ \hline
MetaLDA-dl-def  & Document labels       & Default feature        \\ \hline
MetaLDA-dl-0.01 & Document labels       & Symmetric 0.01 (fixed) \\ \hline
MetaLDA-def-wf  & Default label       & Word features          \\ \hline
MetaLDA-0.1-wf  & Symmetric 0.1 (fixed) & Word features          \\ \hline
MetaLDA-def-def  & Default label & Default feature          \\ \hline
\end{tabular}
}
\end{table}

\subsection{Compared Models and Parameter Settings} We evaluate the performance of the following models:

\begin{itemize}
  \item \textbf{MetaLDA} and its variants: the proposed model and its variants. 
  Here we use MetaLDA to indicate the model considering both document labels and word features. 
  Several variants of MetaLDA with document labels and word features separately were also studied, 
  which are shown in Table~\ref{table-variant}.
  These variants differ in the method of estimating $\vec\alpha$ and $\vec\beta$.
  All the models listed in Table~\ref{table-variant} were implemented on top of Mallet. 
  The hyper-parameters $\mu_0$ and $\nu_0$ were set to $1.0$.
  \item \textbf{LDA}~\cite{blei2003latent}: the baseline model. The Mallet implementation of SparseLDA \cite{yao2009efficient} is used. 
  \item \textbf{LLDA}, Labelled LDA~\cite{ramage2009labeled} and \textbf{PLLDA}, Partially Labelled LDA~\cite{ramage2011partially}: two models that make use of multiple document labels. The original implementation\footnote{\url{https://nlp.stanford.edu/software/tmt/tmt-0.4/}} is used. 
  \item \textbf{DMR}, LDA with Dirichlet Multinomial Regression~\cite{mimno2012topic}: 
  a model that can use multiple document labels. 
  The Mallet implementation of DMR based on SparseLDA was used. 
  Following Mallet, we set the mean of $\lambda$ to 0.0 and set the variances of $\lambda$ for the default label and the document labels to 100.0 and 1.0 respectively. 
  \item \textbf{WF-LDA}, Word Feature LDA~\cite{petterson2010word}: a model with word features. We implemented it on top of Mallet and used the default settings in Mallet for the optimisation. 
  \item \textbf{LF-LDA}, Latent Feature LDA~\cite{nguyen2015improving}: a model that incorporates word embeddings. The original implementation\footnote{\url{https://github.com/datquocnguyen/LFTM}} was used. Following the paper, we used 1500 and 500 MCMC iterations for initialisation and sampling respectively and set $\lambda$ to 0.6,
  and used the original 50-dimensional GloVe word embeddings as word features.
  \item \textbf{GPU-DMM}, Generalized P\'{o}lya Urn DMM~\cite{li2016topic}: a model that incorporates word semantic similarity. The original implementation\footnote{\url{https://github.com/NobodyWHU/GPUDMM}} was used. The word similarity was generated from the distances of the word embeddings. Following the paper, we set the hyper-parameters $\mu$ and $\epsilon$ to 0.1 and 0.7 respectively, and the symmetric document Dirichlet prior to $50/K$. 
  \item \textbf{PTM}, Pseudo document based Topic Model~\cite{zuo2016topic}: a model for short text analysis. The original implementation\footnote{\url{http://ipv6.nlsde.buaa.edu.cn/zuoyuan/}} was used. Following the paper, we set the number of pseudo documents to 1000 and $\lambda$ to 0.1.
\end{itemize}

All the models, except where noted, the symmetric parameters of the document and the topic Dirichlet priors were set to 0.1 and 0.01 respectively,
and 2000 MCMC iterations are used to train the models.

\subsection{Perplexity Evaluation}

Perplexity is  a measure that is widely used~\cite{wallach2009rethinking} to 
evaluate the modelling accuracy of topic models. 
The lower the score, the higher the modelling accuracy.
To compute perplexity, we randomly selected some documents in a dataset as the training set and the remaining as the test set. 
We first trained a topic model on the training set to get the word distributions of each topic $k$ ($\vec{\phi}_k^{train}$). 
Each test document $d$ was split into two halves containing every first and every second words respectively.  
We then fixed the topics and trained the models on the first half to get the topic proportions ($\vec{\theta}_d^{test}$) of test document 
$d$ and compute perplexity for predicting the second half. 
In regard to MetaLDA, we fixed the matrices $\matr{\Phi}^{train}$ and $\matr{\Lambda}^{train}$ output from the training procedure. 
On the first half of test document $d$, 
we computed the Dirichlet prior $\vec{\alpha}_d^{test}$ with $\matr{\Lambda}^{train}$ 
and the labels $\vec{f}_d^{test}$ of test document $d$ (See Step~\ref{step_alpha}),
and then point-estimated $\vec{\theta}_d^{test}$.
We ran all the models 5 times with different random number seeds and report the average scores and the standard deviations.

In testing, 
we may encounter words that never occur in the training documents (a.k.a., unseen words or out-of-vocabulary words). 
There are two strategies for handling unseen words for calculating perplexity on test documents: 
ignoring them or keeping them in computing the perplexity. 
Here we investigate both strategies:

\subsubsection{Perplexity Computed without Unseen Words}

In this experiment, the perplexity is computed only on the words that appear in the training vocabulary. 
Here we used 80\% documents in each dataset as the training set and the remaining 20\% as the test set.

\begin{table*}[]
\centering
\caption{Perplexity comparison on the regular text datasets. The best results are highlighted in boldface.}
\label{table-pp-l}
\centering
\resizebox{\textwidth}{!}{
\begin{tabular}{r|c|c|c|c|c|c|c|c|c|c|c|}

\cline{2-12} 
& Dataset  & \multicolumn{4}{c|}{Reuters} & \multicolumn{4}{c|}{20NG} & \multicolumn{2}{c|}{NYT}  \\ \cline{2-12} 
& \#Topics                      & \textit{50}    & \textit{100}   & \textit{150}   & \textit{200}  & \textit{50}   & \textit{100}  & \textit{150}  & \textit{200}  & \textit{200}  & \textit{500}  \\ \cline{2-12} 
\ldelim \{ {2}{16mm}[\specialcell{No meta info}] 
&LDA                    & 677$\pm$1   & 634$\pm$2   & 629$\pm$1   & 631$\pm$1  & 2147$\pm$7 & 1930$\pm$7 & 1820$\pm$5 & 1762$\pm$3 & 2293$\pm$8 &2154$\pm$4 \\ \cline{2-12}
&MetaLDA-def-def             & 648$\pm$3   & 592$\pm$2   & 559$\pm$1   & 540$\pm$1  & 2093$\pm$6 & 1843$\pm$7 & 1708$\pm$5 & 1626$\pm$4 & 2258$\pm$9 & 2079$\pm$8 \\ \cline{2-12}  
\ldelim \{ {3}{13mm}[\specialcell{Doc labels}]
&DMR    & 640$\pm$1   & 577$\pm$1   & 544$\pm$2   & 526$\pm$2  & 2080$\pm$8 & 1811$\pm$8 & 1670$\pm$4 & 1578$\pm$1  &\textbf{2231}$\pm$13 &\textbf{2013}$\pm6$ \\ \cline{2-12}
&MetaLDA-dl-0.01       & 649$\pm$2   & 582$\pm$2   & 551$\pm$3   & 530$\pm$2  & 2067$\pm$9 & 1821$\pm$7 & 1680$\pm$5 & 1590$\pm$1 & \textbf{2219}$\pm$4 & \textbf{2018}$\pm$4 \\ \cline{2-12}
&MetaLDA-dl-def      & 642$\pm$3   & 576$\pm$3   & 543$\pm$1   & 526$\pm$1  & 2050$\pm$4 & 1804$\pm$6 & 1675$\pm$8 & 1589$\pm$2 & \textbf{2230}$\pm$3 & \textbf{2022}$\pm$5 \\ \cline{2-12}
\ldelim \{ {4}{16.5mm}[\specialcell{Word features}] 
&LF-LDA                 & 841$\pm$4   & 787$\pm$4   & 772$\pm$3   & 771$\pm$4  & 2855$\pm$21 & 2576$\pm$3 & 2433$\pm$7 & 2326$\pm$8 &2831$\pm$2 & 2700$\pm$5 \\ \cline{2-12}
&WF-LDA & 659$\pm$2   & 616$\pm$2   & 615$\pm$1   & 613$\pm$1  & 2089$\pm$7 & 1875$\pm$2 & 1784$\pm$2 & 1727$\pm$3  & 2287$\pm$6 & 2134$\pm$6\\ \cline{2-12}

&MetaLDA-0.1-wf  & 659$\pm$3   & 621$\pm$1   & 619$\pm$1   & 623$\pm$1  & 2098$\pm$7 & 1887$\pm$8 & 1796$\pm$8 & 1744$\pm$4 & 2283$\pm$4 &2143$\pm$2 \\ \cline{2-12}
&MetaLDA-def-wf      & 643$\pm$2   & 582$\pm$4   & 552$\pm$3   & 535$\pm$1  & 2068$\pm$6 & 1819$\pm$1 & 1685$\pm$7 & 1600$\pm$3 & 2260$\pm$7 & 2095$\pm$6  \\ \cline{2-12}

 \specialcell{Doc labels \& \\word features}   $\longrightarrow$ &MetaLDA          & \textbf{633}$\pm$2   & \textbf{568}$\pm$2   & \textbf{536}$\pm$2   & \textbf{517}$\pm$1  & \textbf{2025}$\pm$12 & \textbf{1781}$\pm$8 & \textbf{1640}$\pm$5 & \textbf{1551}$\pm$6 & \textbf{2217}$\pm$6 & \textbf{2020}$\pm$6\\ \cline{2-12}
\end{tabular}
}
\centering

\begin{tabular}{r|c|c|c|c|c|c|c|c|c|c|c|}
\cline{2-12} 
& Dataset    & \multicolumn{4}{c|}{Reuters} & \multicolumn{4}{c|}{20NG} & \multicolumn{2}{c|}{NYT} \\ \cline{2-12} 
&\specialcell{\#Topics per label}  & \textit{5}    & \textit{10}   & \textit{20}   & \textit{50}  & \textit{5}   & \textit{10}  & \textit{20}  & \textit{50} & \textit{2}  & \textit{5}  \\ \cline{2-12} 
\ldelim \{ {2}{13mm}[\specialcell{Doc labels}]&   PLLDA                  & 714   & 708   & 733   &  829    & 1997 & 1786 & 1605 & \textbf{1482} & 2839 &  2846  \\ \cline{2-12}
& LLDA                   & \multicolumn{4}{c|}{834}     & \multicolumn{4}{c|}{2607}  & \multicolumn{2}{c|}{2948} \\ \cline{2-12} 
\end{tabular}
\end{table*}

\begin{table*}
\centering
\caption{Perplexity comparison without unseen words on the short text datasets. The best results are highlighted in boldface.}
\label{table-pp-s}
\centering
\resizebox{\textwidth}{!}{

\begin{tabular}{r|c|c|c|c|c|c|c|c|c|c|c|}
\cline{2-12} 
& Dataset       & \multicolumn{4}{c|}{WS}   & \multicolumn{4}{c|}{TMN}  & \multicolumn{2}{c|}{AN} \\ \cline{2-12} 
&\#Topics                        & \textit{50}    & \textit{100}  & \textit{150} & \textit{200}  & \textit{50}   & \textit{100}  & \textit{150}  & \textit{200}  & \textit{50}   & \textit{100}   \\ \cline{2-12}
\ldelim \{ {2}{16mm}[\specialcell{No meta info}] &LDA                     & 961$\pm$6   & 878$\pm$8  & 869$\pm$6 & 888$\pm$5  & 1969$\pm$14   & 1873$\pm$6  & 1881$\pm$9 & 1916$\pm$4  & 406$\pm$14  & 422$\pm$12   \\ \cline{2-12}
&MetaLDA-def-def               & 884$\pm$10   & 733$\pm$6  & 671$\pm$6 & 625$\pm$6  & 1800$\pm$11 & 1578$\pm$19 & 1469$\pm$4 & 1422$\pm$6 & 352$\pm$16  & 336$\pm$11    \\  \cline{2-12} 
\ldelim \{ {3}{13mm}[\specialcell{Doc labels}]&DMR                     & 845$\pm$7   & 683$\pm$4  & 607$\pm$1 & 562$\pm$2  & 1750$\pm$8 & 1506$\pm$3 & 1391$\pm$7 & 1323$\pm$5 & 326$\pm$6  & \textbf{290}$\pm$5  \\ \cline{2-12} 
&MetaLDA-dl-0.01                     & 840$\pm$7   & 693$\pm$6  & 618$\pm$3 & 588$\pm$4  & 1767$\pm$11 & 1528$\pm$10 & 1416$\pm$7 & 1345$\pm$13 & 321$\pm$13  & 303$\pm$8  \\ \cline{2-12} 
&MetaLDA-dl-def                     & 832$\pm$4   & 679$\pm$5  & 622$\pm$7 & 582$\pm$5  & 1720$\pm$7 & 1505$\pm$16 & 1395$\pm$11 & 1325$\pm$12 & 319$\pm$9  & 293$\pm$7  \\ \cline{2-12} 

\ldelim \{ {4}{16.5mm}[\specialcell{Word features}]&LF-LDA                 & 1164$\pm$6  & 1039$\pm$17 & 1019$\pm$11 & 992$\pm$6 & 2415$\pm$35 & 2393$\pm$11 & 2371$\pm$10 & 2374$\pm$14 & 482$\pm$17  & 514$\pm$19  \\  \cline{2-12} 
&WF-LDA                  & 894$\pm$6   & 839$\pm$6  & 827$\pm$10 & 842$\pm$4  & 1853$\pm$6  & 1766$\pm$12 & 1830$\pm$60   & 1854$\pm$45   & 397$\pm$5  & 410$\pm$6   \\  \cline{2-12} 
&MetaLDA-0.1-wf                  & 889$\pm$6   & 832$\pm$3  & 839$\pm$2 & 853$\pm$4  & 1865$\pm$4  & 1784$\pm$2 & 1799$\pm$9   & 1831$\pm$6   & 388$\pm$3  & 410$\pm$8   \\  \cline{2-12} 
&MetaLDA-def-wf                  & 830$\pm$6   & 688$\pm$8  & 624$\pm$5 & 584$\pm$4  & 1730$\pm$14  & 1504$\pm$3 & 1402$\pm$13   & 1342$\pm$4   & 346$\pm$15  & 332$\pm$8   \\  \cline{2-12}

\specialcell{Doc labels \& \\word features}   $\longrightarrow$ &MetaLDA      & \textbf{774}$\pm$9   & \textbf{627}$\pm$6  & \textbf{572}$\pm$3 & \textbf{534}$\pm$4  & \textbf{1657}$\pm$4 & \textbf{1415}$\pm$16 & \textbf{1304}$\pm$6 & \textbf{1235}$\pm$6 & \textbf{314}$\pm$9  & \textbf{293}$\pm$9  \\\cline{2-12} 
\end{tabular}

}
\centering
\begin{tabular}{r|c|c|c|c|c|c|c|c|c|c|c|}
\cline{2-12} 
& Dataset       & \multicolumn{4}{c|}{WS}   & \multicolumn{4}{c|}{TMN}  & \multicolumn{2}{c|}{AN} \\ \cline{2-12} 

&\#Topics per label  & \textit{5}    & \textit{10}  & \textit{20} & \textit{50}  & \textit{5}   & \textit{10}  & \textit{20}  & \textit{50}  & \textit{5}   & \textit{10} \\ \cline{2-12} 
\ldelim \{ {2}{13mm}[\specialcell{Doc labels}]&PLLDA                & 1060  & 886  & 735 & 642  & 2181 & 1863 & 1647 & 1456 &     440 &  525     \\ \cline{2-12}
&LLDA                   & \multicolumn{4}{c|}{1543} & \multicolumn{4}{c|}{2958} & \multicolumn{2}{c|}{392}   \\\cline{2-12}
\end{tabular}

\end{table*}

Tables~\ref{table-pp-l} and \ref{table-pp-s} show\footnote{For GPU-DMM and PTM, 
perplexity is not evaluated because the inference code for unseen 
documents is not public available. The random number seeds used in 
the code of LLDA and PLLDA are pre-fixed in the package.
So the standard deviations of the two models are not reported.}: 
the average perplexity scores with standard 
deviations for all the models. Note that: (1) The scores on AN with 150 and 200 topics are not reported
due to overfitting observed in all the compared models. (2) Given the size of NYT, the scores of 200 and 500 topics
are reported. (3) The number of latent topics in LLDA must equal to the number of document labels. 
(4) For PLLDA, we varied the number of topics per label from 5 to 50 (2 and 5 topics on NYT).
The number of topics in PPLDA is the product of the numbers of labels and topics per label.


\begin{figure*}[t]
        \centering
        \caption{Perplexity comparison with unseen words in different proportions of the training documents. Each pair of the numbers on the horizontal axis are the proportion of the training documents and the proportion of unseen tokens in the vocabulary of the test documents, respectively. The error bars are the standard deviations over 5 runs.}
        \label{fig-pp-unseen}
         \begin{subfigure}[b]{0.244\linewidth}
                 \centering
                 \caption{Reuters with 200 topics}
                 \includegraphics[width=1.0\textwidth]{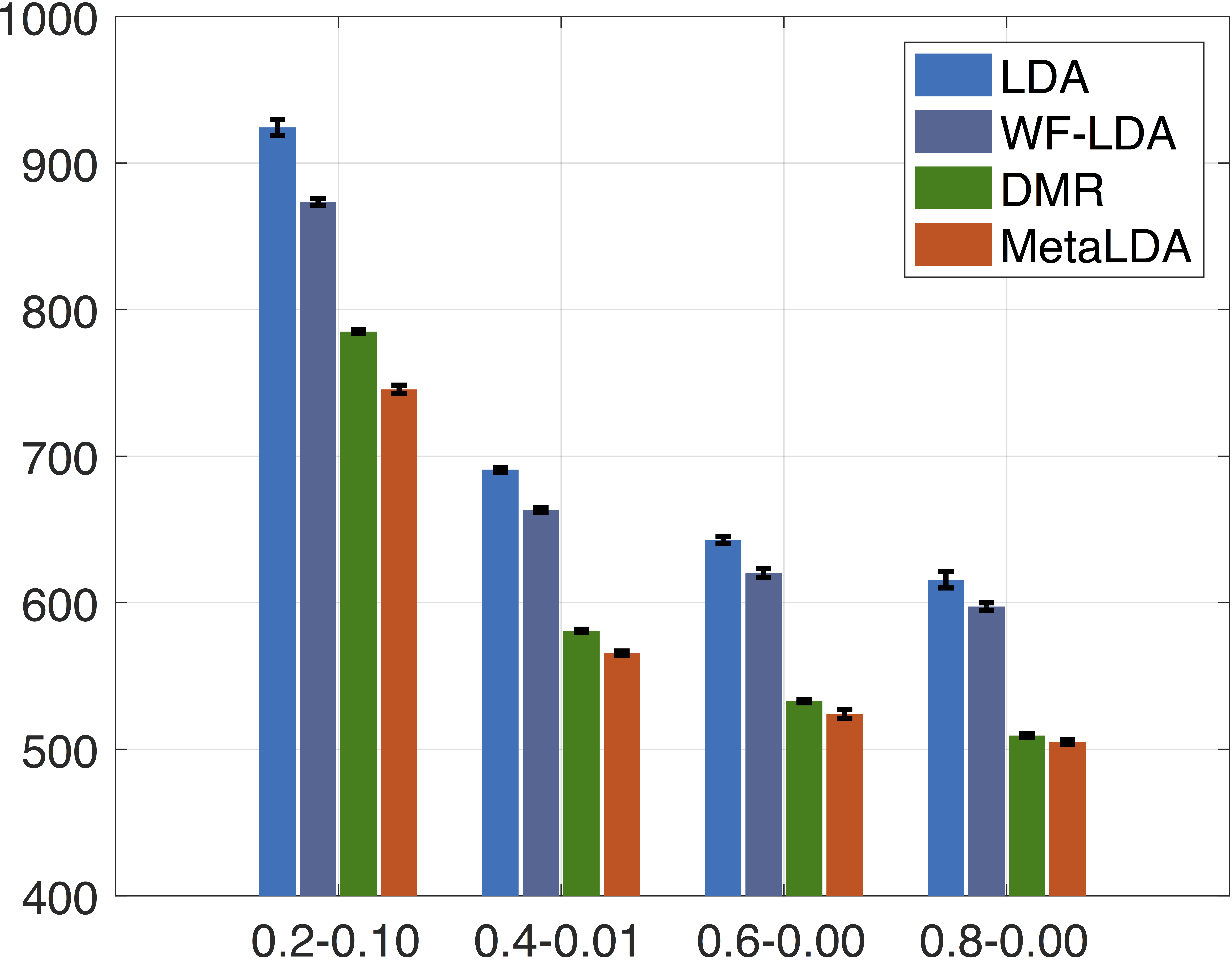}
         \end{subfigure}%
              \begin{subfigure}[b]{0.244\linewidth}
                 \centering
                 \caption{20NG with 200 topics}
                 \includegraphics[width=1.0\textwidth]{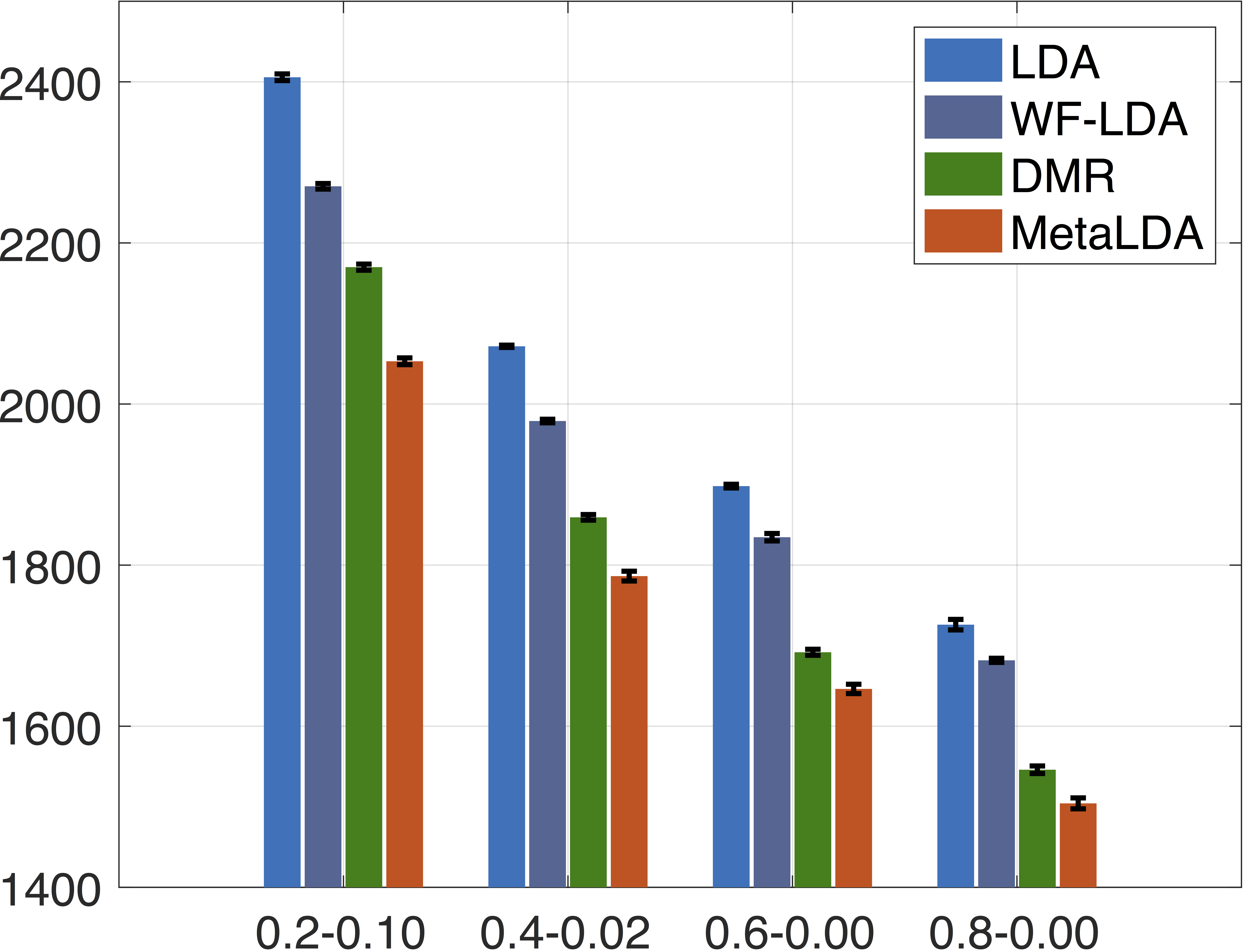}
         \end{subfigure}
                 \begin{subfigure}[b]{0.244\linewidth}
                 \centering
                 \caption{TMN with 100 topics}
                 \includegraphics[width=1.0\textwidth]{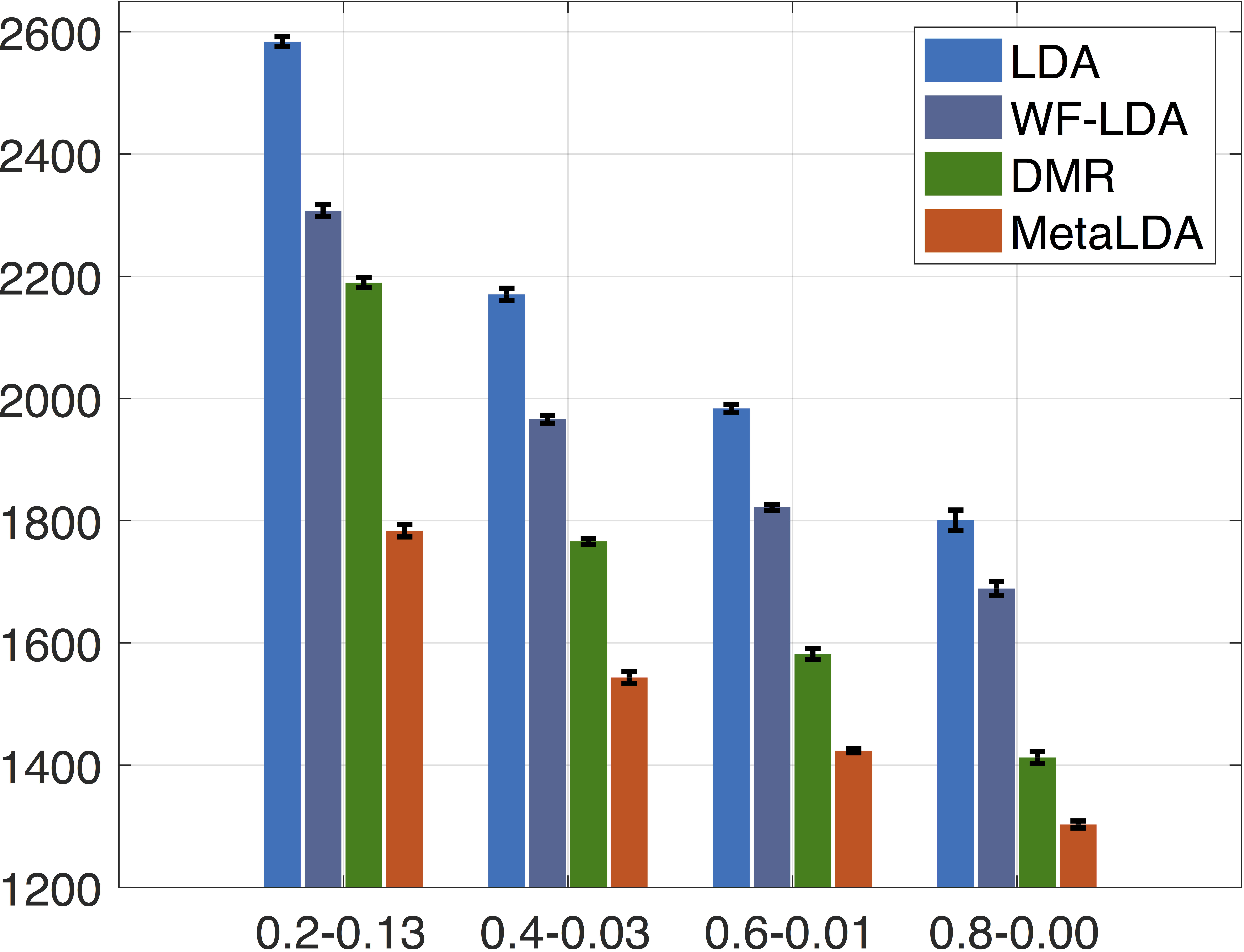}
         \end{subfigure}
            \begin{subfigure}[b]{0.244\linewidth}
                 \centering
                 \caption{WS with 50 topics}
                 \includegraphics[width=1.0\textwidth]{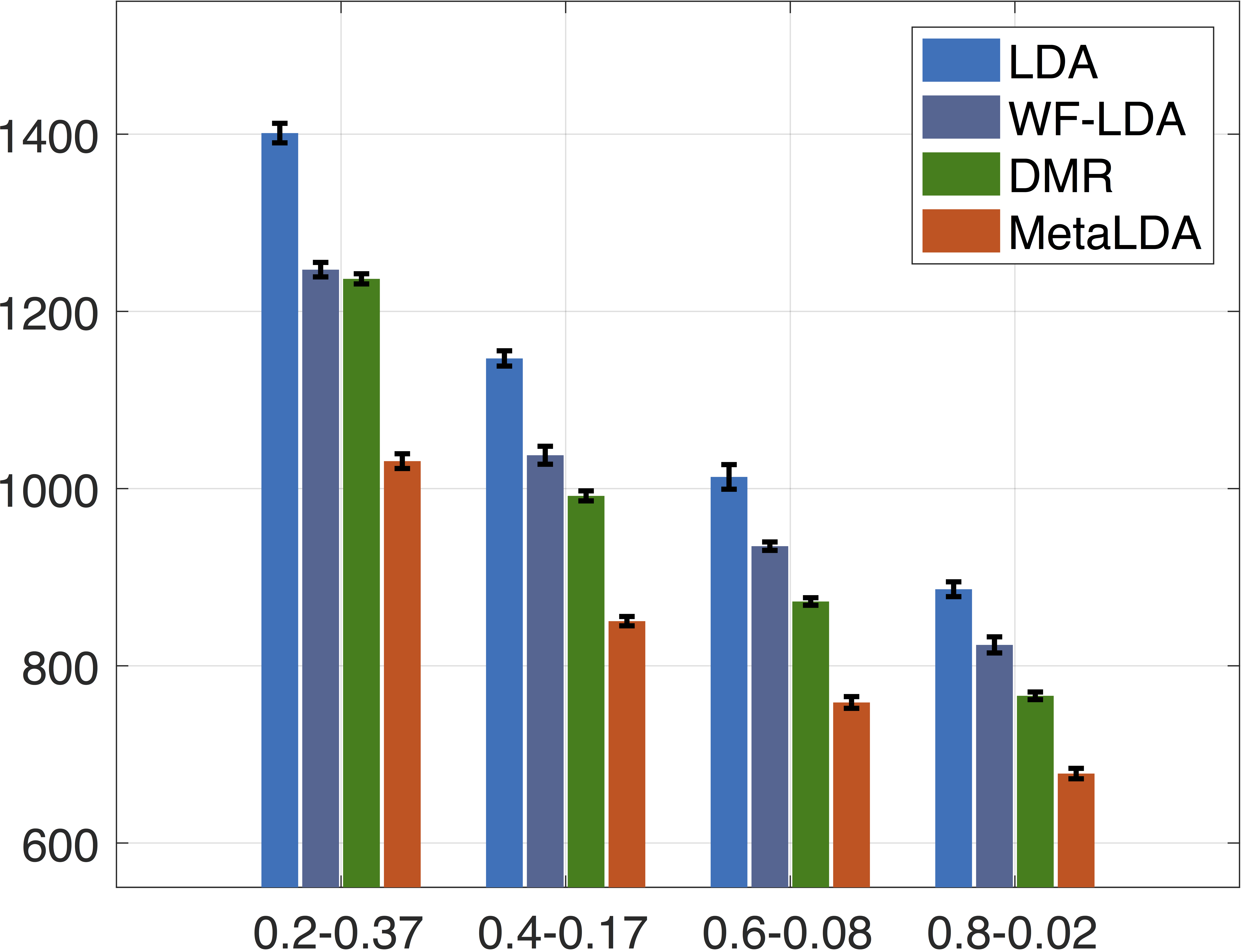}
         \end{subfigure}
     \end{figure*} 
 
The results show that MetaLDA outperformed all the competitors 
in terms of perplexity on nearly all the datasets, 
showing the benefit of using both document and word meta information. 
Specifically, we have the following remarks:
\begin{itemize}
\item By looking at the models using only the document-level meta information, 
we can see the significant improvement of these models over LDA, 
which indicates that document labels can play an important role in guiding topic modelling. 
Although the performance of the two variants of MetaLDA with document labels and DMR is comparable, 
our models runs much faster than DMR, which will be studied later in Section~\ref{subsection-speed}.
\item It is interesting that PLLDA with 50 topics for each label has better perplexity 
than MetaLDA with 200 topics in the 20NG dataset. 
With the 20 unique labels, the actual number of topics in PLLDA is 1000. 
However, 
if 10 topics for each label in PLLDA are used, which is equivalent to 200 topics in MetaLDA, 
PLLDA is outperformed by MetaLDA significantly.
\item At the word level, MetaLDA-def-wf performed the best among the models with word features only. Moreover, our model has obvious advantage in running speed~(see Table~\ref{table-time}). 
Furthermore, comparing MetaLDA-def-wf with MetaLDA-def-def and MetaLDA-0.1-wf with LDA, we can see using the word features indeed improved perplexity.
\item The scores show that the improvement gained by MetaLDA over LDA 
on the short text datasets is larger than that on the regular text datasets. 
This is as expected because meta information serves as complementary information 
in MetaLDA and can have more significant impact when the data is sparser.
\item It can be observed that models usually gained improved perplexity, if $\alpha$ is sampled/optimised,
in line with \cite{wallach2009rethinking}.
\item On the AN dataset, there is no statistically significant difference between MetaLDA and DMR.
On NYT, a similar trend is observed: the improvement in the models with the document labels over LDA is obvious but not in the models with the word features. 
Given the number of the document labels (194 of AN and 545 of NYT), 
it is possible that the document labels already offer enough information and the word embeddings 
have little contribution in the two datasets.  
\end{itemize}

\subsubsection{Perplexity Computed with Unseen Words}
To test the hypothesis that the incorporation of meta information in MetaLDA
can significantly improve the modelling accuracy in the cases where the
corpus is sparse, we varied the proportion of documents used in training 
from 20\% to 80\% and used the remaining for testing.
It is natural that when the proportion is small, the number of unseen 
words in testing documents will be large.
Instead of simply excluding the unseen words in the previous experiments, here we compute the perplexity with unseen words for LDA, DMR, WF-LDA and the proposed MetaLDA. For perplexity calculation, $\phi^{test}_{k,v}$ for each topic $k$ and each token $v$ in the test documents is needed. 
If $v$ occurs in the training documents, $\phi^{test}_{k,v}$ can be directly obtained. While if $v$ is unseen, 
$\phi^{unseen}_{k,v}$ can be estimated by the prior: 
$\frac{\beta^{unseen}_{k,v}}{n^{train}_{k,\cdot} + \beta^{train}_{k,\cdot} + \beta^{unseen}_{k,\cdot}}$. 
For LDA and DMR which do not use word features, $\beta^{unseen}_{k,v} = \beta^{train}_{k,v}$; 
For WF-LDA and MetaLDA which are with word features, $\beta^{unseen}_{k,v}$ is computed with the features of the unseen token. Following Step~\ref{step_beta}, for MetaLDA, 
$\beta^{unseen}_{k,v} = \prod_{l'}^{L_{word}} \delta_{l',k}^{g^{unseen}_{v,l}}$.



Figure~\ref{fig-pp-unseen} shows the perplexity scores on Reuters, 20NG, TMN and WS with 
200, 200, 100 and 50 topics respectively. 
MetaLDA outperformed the other models significantly with 
a lower proportion of training documents and relatively higher proportion of unseen words. 
The gap between MetaLDA and the other three models increases while the training proportion
decreases.
It indicates that the meta information helps MetaLDA to achieve 
better modelling accuracy on predicting unseen words. 

\subsection{Topic Coherence Evaluation}
We further evaluate the semantic coherence of the words in a topic learnt by LDA, PTM, DMR,
LF-LDA, WF-LDA, GPU-DMM and MetaLDA. 
Here we use the Normalised Pointwise Mutual Information (NPMI)~\cite{aletras2013evaluating,lau2014machine} 
to calculate topic coherence score for topic $k$ with top $T$ words: 
$\text{NPMI}(k) = \sum_{j=2}^{T} \sum_{i = 1}^{j-1} \log\frac{p(w_{j},w_{i})}{p(w_j)p(w_i)}/-\log p(w_j, w_i)$,
where $p(w_i)$ is the probability of word $i$, 
and $p(w_i,w_j)$ is the joint probability of words $i$ 
and $j$ that co-occur together within a sliding window. 
Those probabilities were computed on an external large corpus, i.e., a 5.48GB Wikipedia dump in our experiments. 
The NPMI score of each topic in the experiments is calculated with top 10 words ($T=10$) 
by the Palmetto package\footnote{\url{http://palmetto.aksw.org}}. 
Again, we report the average scores and the standard deviations over 5 random runs.


\begin{table*}[!htb]
\caption{Topic coherence (NPMI) on the short text datasets.}
\centering
\label{table-npmi}
\begin{tabular}{r|c|c|c|c|c|c|c|}
\cline{2-8}
&&       \multicolumn{3}{c|}{All 100 topics} & \multicolumn{3}{c|}{Top 20 topics} \\ \cline{2-8}
 &       & WS & TMN & AN  & WS & TMN & AN  \\ \cline{2-8}
\ldelim \{ {2}{16mm}[\specialcell{No meta info}]&LDA     & -0.0030$\pm$0.0047 & 0.0319$\pm$0.0032   & -0.0636$\pm$0.0033   &  0.1025$\pm$0.0067 & 0.137$\pm$0.0043   & -0.0010$\pm$0.0052  \\ \cline{2-8}
&PTM     & -0.0029$\pm$0.0048  &  0.0355$\pm$0.0016 &    -0.0640$\pm$0.0037   &  0.1033$\pm$0.0081 &  0.1527$\pm$0.0052 &    0.0004$\pm$0.0037    \\ \cline{2-8}
 \specialcell{Doc labels}   $\rightarrow$  &DMR     & 0.0091$\pm$0.0046 & 0.0396$\pm$0.0044   & -0.0457$\pm$0.0024   &  0.1296$\pm$0.0085 & 0.1472$\pm$0.1507   & 0.0276$\pm$0.0101  \\ \cline{2-8}
\ldelim \{ {3}{17mm}[\specialcell{Word features}]&LF-LDA  & 0.0130$\pm$0.0052  & 0.0397$\pm$0.0026 &   -0.0523$\pm$0.0023 &  0.1230$\pm$0.0153  & 0.1456$\pm$0.0087  &  0.0272$\pm$0.0042  \\ \cline{2-8}
&WF-LDA  & 0.0091$\pm$0.0046  & 0.0390$\pm$0.0051  &  -0.0457$\pm$0.0024  & 0.1296$\pm$0.0085  & 0.1507$\pm$0.0055  &  0.0276$\pm$0.0101   \\ \cline{2-8}
&GPU-DMM & -0.0934$\pm$0.0106 & -0.0970$\pm$0.0034   & -0.0769$\pm$0.0012   &  0.0836$\pm$0.0105 &  0.0968$\pm$0.0076   &  -0.0613$\pm$0.0020     \\ \cline{2-8}
 \specialcell{Doc labels \& \\word features}   $\rightarrow$  &MetaLDA  & \textbf{0.0311}$\pm$0.0038  &  \textbf{0.0451}$\pm$0.0034  &  \textbf{-0.0326}$\pm$0.0019   & \textbf{0.1511}$\pm$0.0093  &     \textbf{0.1584}$\pm$0.0072  &  \textbf{0.0590}$\pm$0.0065     \\ \cline{2-8}
\end{tabular}
\end{table*}

\begin{table*}[]
\centering
\caption{Running time (seconds per iteration) on 80\% documents of each dataset.}
\label{table-time}
\begin{tabular}{r|c|c|c|c|c|c|c|c|c||c|c|}
\cline{2-12}
                  & Dataset  & \multicolumn{4}{c|}{Reuters} & \multicolumn{4}{c||}{WS} & \multicolumn{2}{c|}{NYT} \\ \cline{2-12} 
                  & \#Topics & 50   & 100  & 150  & 200  & 50  & 100  & 150  & 200 & 200 & 500 \\ \cline{2-12} 
\ldelim \{ {2}{16mm}[\specialcell{No meta info}]                   
& LDA      &  0.0899    &  0.1023    & 0.1172     &  0.1156    &  0.0219   &   0.0283   &  0.0301    &  0.0351  & 0.7509 &  1.1400 \\ \cline{2-12} 
                  & PTM      &  4.9232
    &  5.8885    &  7.2226    & 7.7670     &  1.1840   &  1.6375    &   1.8288   &  2.0030 & - & -  \\ \cline{2-12} 
\ldelim \{ {2}{13mm}[\specialcell{Doc labels}]                  
& DMR      &    0.6112   &   0.9237   &  1.2638    &  1.6066    &  0.4603   &  0.8549    &  1.2521    &  1.7173 & 13.7546 & 31.9571   \\ \cline{2-12} 
& MetaLDA-dl-0.01      &    0.1187   &   0.1387   &  0.1646    &  0.1868     &  0.0396   &  0.0587    &  0.0769    & 0.112 1  & 2.4679 &  4.9928 \\ \cline{2-12} 
\ldelim \{ {4}{17mm}[\specialcell{Word features}] 
  & LF-LDA   &  2.6895    & 5.3043     &  8.3429    &   11.4419   &  2.4920   &   6.0266   & 9.1245     &  11.5983 & 95.5295 & 328.0862 \\\cline{2-12}
  & WF-LDA   &  1.0495    & 1.6025     &  3.0304    &  4.8783    &  1.8162   & 3.7802     &  6.1863    &  8.6599   & 14.0538 & 31.4438 \\ \cline{2-12} 
  & GPU-DMM   &  0.4193    & 0.7190     &  1.0421    &  1.3229    &  0.1206   & 0.1855      &  0.2487    &  0.3118 & - & -  \\ \cline{2-12}
  & MetaLDA-0.1-wf   &  0.2427    & 0.4274     &  0.6566    &  0.9683    &   0.1083   & 0.1811      &  0.2644    &  0.3579 & 4.6205 & 12.4177 \\ \cline{2-12} 
 \specialcell{Doc labels \& \\word features}   $\rightarrow$                 
 & MetaLDA  &  0.2833    &  0.5447    &  0.7222    &  1.0615    &  0.1232   &  0.2040    &  0.3282    & 0.4167 & 6.4644 & 16.9735   \\ \cline{2-12} 
\end{tabular}
\end{table*}

It is known that conventional topic models directly applied to short texts
suffer from low quality topics, caused by the insufficient word co-occurrence 
information.
Here we study whether or not the meta information helps MetaLDA improve topic quality, 
compared with other topic models that can also handle short texts. 
Table~\ref{table-npmi} shows the NPMI scores on the three short text datasets.
Higher scores indicate better topic coherence.
All the models were trained with 100 topics. 
Besides the NPMI scores averaged over all the 100 topics, 
we also show the scores averaged over top 20 topics with highest NPMI,
where ``rubbish'' topics are eliminated, following \cite{yang2015efficient}.
It is clear that MetaLDA performed significantly better than all the other models in WS and AN dataset in terms of NPMI,
which indicates that MetaLDA can discover more meaningful topics with the document and word meta information.
We would like to point out that on the TMN dataset, even though the average score of MetaLDA is still the best, 
the score of MetaLDA has overlapping with the others' in the standard deviation,
which indicates the difference is not statistically significant.

\subsection{Running Time}
\label{subsection-speed}
In this section, we empirically study the efficiency of the models in term of per-iteration running time.
The implementation details of our MetaLDA are as follows:
(1) The SparseLDA framework \cite{yao2009efficient} reduces the complexity of LDA to be sub-linear 
by breaking the conditional of LDA into three ``buckets'', where the ``smoothing only'' bucket is cached 
for all the documents and the ``document only'' bucket is cached for all the tokens in a document. 
We adopted a similar strategy when implementing MetaLDA.
When only the document meta information is used, 
the Dirichlet parameters $\alpha$ for different documents in MetaLDA are different and asymmetric. 
Therefore, the ``smoothing only'' bucket has to be computed for each document, 
but we can cache it for all the tokens, which still gives us a considerable reduction in 
computing complexity.
However, when the word meta information is used, 
the SparseLDA framework no longer works in MetaLDA as the $\beta$ parameters
for each topic and each token are different. 
(2) By adapting the DistributedLDA framework~\cite{newman2009distributed}, 
our MetaLDA implementation runs in parallel with multiple threads, 
which makes MetaLDA able to handle larger document collections.
The parallel implementation was used on the NYT dataset.


The per-iteration running time of all the models is shown in Table~\ref{table-time}. 
Note that: (1) On the Reuters and WS datasets, all the models ran with a single thread
on a desktop PC with a 3.40GHz CPU and 16GB RAM. 
(2) Due to the size of NYT, we report the running time for the models that are able to run in parallel. All the parallelised models ran with 10 threads on a cluster with a 14-core 2.6GHz CPU and 128GB RAM. (3) All the models were implemented in JAVA.
(4) As the models with meta information add extra complexity to LDA, the per-iteration running time of LDA can be treated as the lower bound. 

At the document level, both MetaLDA-df-0.01 and DMR use priors to incorporate the document meta information
and both of them were implemented in the SparseLDA framework. However,
our variant is about 6 to 8 times faster than DMR on the Reuters dataset and more than 10 times faster on the WS dataset. 
Moreover, it can be seen that the larger the number of topics, the faster our variant is over
DMR.
At the word level, similar patterns can be observed:
our MetaLDA-0.1-wf ran significantly faster than WF-LDA and LF-LDA especially when more topics are used (20-30 times faster on WS).
It is not surprising that GPU-DMM has comparable running speed with our variant, because 
only one topic is allowed for each document in GPU-DMM. 
With both document and word meta information, MetaLDA still ran several times faster than DMR, 
LF-LDA, and WF-LDA. On NYT with the parallel settings, MetaLDA maintains its efficiency advantage as well.




\section{Conclusion}

In this paper, we have presented a topic modelling framework named MetaLDA that 
can efficiently incorporate document and word meta information.
This gains 
a significant improvement over others in terms of perplexity and topic quality. 
With two data augmentation techniques, 
MetaLDA enjoys full local conjugacy, allowing efficient Gibbs sampling,
demonstrated by superiority in the per-iteration running time.
Furthermore, without losing generality, MetaLDA can work with both regular texts and short texts. 
The improvement of MetaLDA over other models that also use meta information 
is more remarkable, 
particularly when the word-occurrence information is insufficient.
As MetaLDA takes a particular approach
for incorporating meta information on topic models, 
it is possible to apply the same approach
to other Bayesian probabilistic models, where Dirichlet priors are used.
Moreover, it would be interesting to extend our method to use real-valued meta information directly, 
which is the subject of future work.




\section*{Acknowledgement}

Lan Du was partially supported by Chinese NSFC project under grant number 61402312.
Gang Liu was partially supported by Chinese PostDoc Fund under grant number LBH-Q15031.

\bibliographystyle{IEEEtran}
\bibliography{IEEEabrv,icdm}

\end{document}

%% file: ExtLDA_graphical_model.tex
%
%
%
%


\begin{tikzpicture}[x=1cm,y=1cm]


  \node[latent]    (z)      {$z_{d,i}$} ; %

  \node[obs, below=of z]                   (w)      {$w_{d,i}$} ; %
  \node[latent, left=of z]    (theta)  {$\vec{\theta_d}$}; %
  \node[latent, left=of theta] (alpha) {$\vec{\alpha_d}$};

  \node[obs,  below=of alpha] (f) {$f_{d,l}$};

  \node[latent, left=of f, yshift=0.3cm] (lambda) {$\lambda_{l,k}$};

  \node[const, above=of lambda] (mu_0) {$\mu_0$};



  \node[latent, right=of w] (phi)  {$\vec{\phi_k}$}; %
  \node[latent, above=of phi]  (beta) {$\vec{\beta_k}$}; %

  \node[latent, above=of beta] (delta) {$\delta_{l',k}$};
  \node[obs,  above=of z, yshift=0.4cm] (g) {$g_{v,l'}$};
  \node[const, above=of delta] (nu_0) {$\nu_0$};


  \edge {theta}      {z} ; %
  \edge {alpha}  {theta} ; %
  \edge {phi}        {w} ; %
  \edge {beta}    {phi} ; %
  \edge {z} {w};
  \edge {f} {alpha};
  \edge {lambda} {alpha};
  \edge {mu_0} {lambda};
  \edge {g} {beta};
  \edge {delta} {beta};
  \edge {nu_0} {delta};

  \plate {lambda-k}
  {
    (lambda)
  } {$\forall~k$};
  \plate {f-l}
  {
    (f)
    (lambda-k)
  } {$\forall~l~~~~~~~~~~~~~~~~~$};


  \plate {g-v}
  {
    (g)
  } {$\forall~v$};

  \plate[xshift=0.05cm] {g-l}
  {
    (g-v)
    (delta)
  } {$\forall~l'~~~~~~~~~~~~~~~$};

  \plate {word} { %
    (z)
    (w)
  } {$\forall~i$}; %
  \plate {} { %
    (word) %
    (f) (alpha)(theta) %
  } {$\forall~d$} ; %
  \plate {} { %
    (phi) %
    (beta)
    (delta)
  } {$\forall~k$} ; %

\end{tikzpicture}
